\documentclass[final]{cvpr}

\usepackage{times}
\usepackage{epsfig}
\usepackage{graphicx}
\usepackage{amsmath}
\usepackage{amssymb}

\usepackage[utf8]{inputenc} 
\usepackage[T1]{fontenc}    

\usepackage[dvipsnames]{xcolor}
\usepackage{url}            
\usepackage{booktabs}       
\usepackage{amsfonts}       
\usepackage{nicefrac}       
\usepackage{microtype}      

\usepackage{subfigure}
\usepackage{booktabs} 
\usepackage{xcolor}
\usepackage{multirow}
\usepackage{pifont}

\usepackage{algorithm}
\usepackage[noend]{algpseudocode}
\usepackage{setspace}
\usepackage{stfloats}
\usepackage{marvosym}

\newcommand{\myparagraph}[1]{\vspace{-8pt}\paragraph{#1}}

\usepackage[pagebackref=true,breaklinks=true,colorlinks,bookmarks=false]{hyperref}

\def\cvprPaperID{2771}
\def\confYear{CVPR 2021}
\begin{document}

\title{MetaSets: Meta-Learning on Point Sets for Generalizable Representations}

\author{Chao Huang\thanks{Equal contribution} ,  Zhangjie Cao\footnotemark[1] , Yunbo Wang\footnotemark[1] , Jianmin Wang, Mingsheng Long (\Letter) \\
School of Software, BNRist, Tsinghua University, China \\
{\tt\small \{huangcthu,caozhangjie14,yunbo.thu\}@gmail.com, \{jimwang,mingsheng\}@tsinghua.edu.cn}
\and
}

\maketitle

\begin{abstract}
Deep learning techniques for point clouds have achieved strong performance on a range of 3D vision tasks. However, it is costly to annotate large-scale point sets, making it critical to learn generalizable representations that can transfer well across different point sets. In this paper, we study a new problem of 3D Domain Generalization (3DDG) with the goal to generalize the model to other unseen domains of point clouds without any access to them in the training process. It is a challenging problem due to the substantial geometry shift from simulated to real data, such that most existing 3D models underperform due to overfitting the complete geometries in the source domain. We propose to tackle this problem via MetaSets, which meta-learns point cloud representations from a group of classification tasks on carefully-designed transformed point sets containing specific geometry priors. The learned representations are more generalizable to various unseen domains of different geometries. We design two benchmarks for Sim-to-Real transfer of 3D point clouds. Experimental results show that MetaSets outperforms existing 3D deep learning methods by large margins.
\end{abstract}
\section{Introduction}

\begin{figure}[ht]
    \centering
    {\includegraphics[width=0.98\columnwidth]{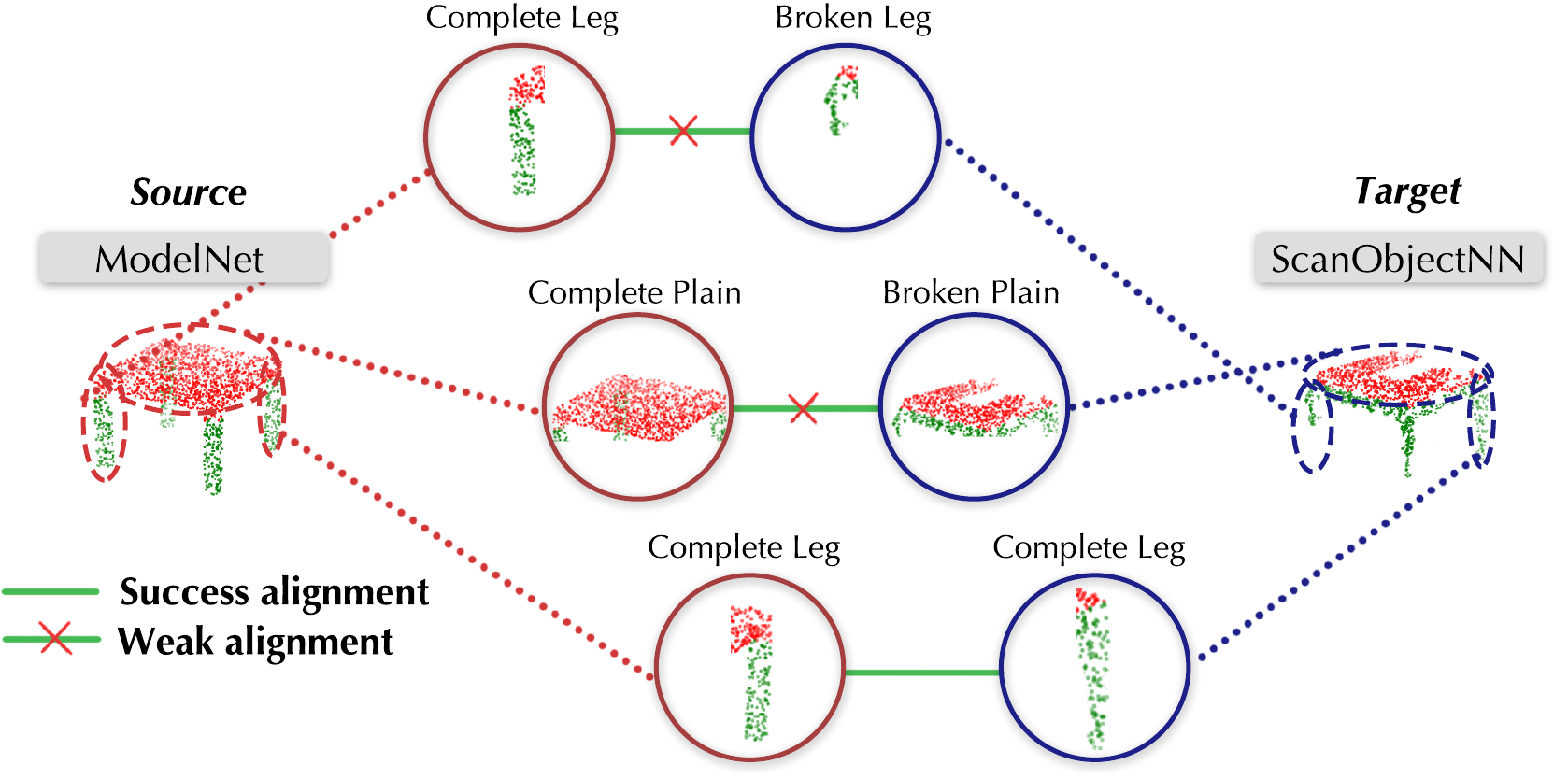}}
    {\includegraphics[width=.8\columnwidth]{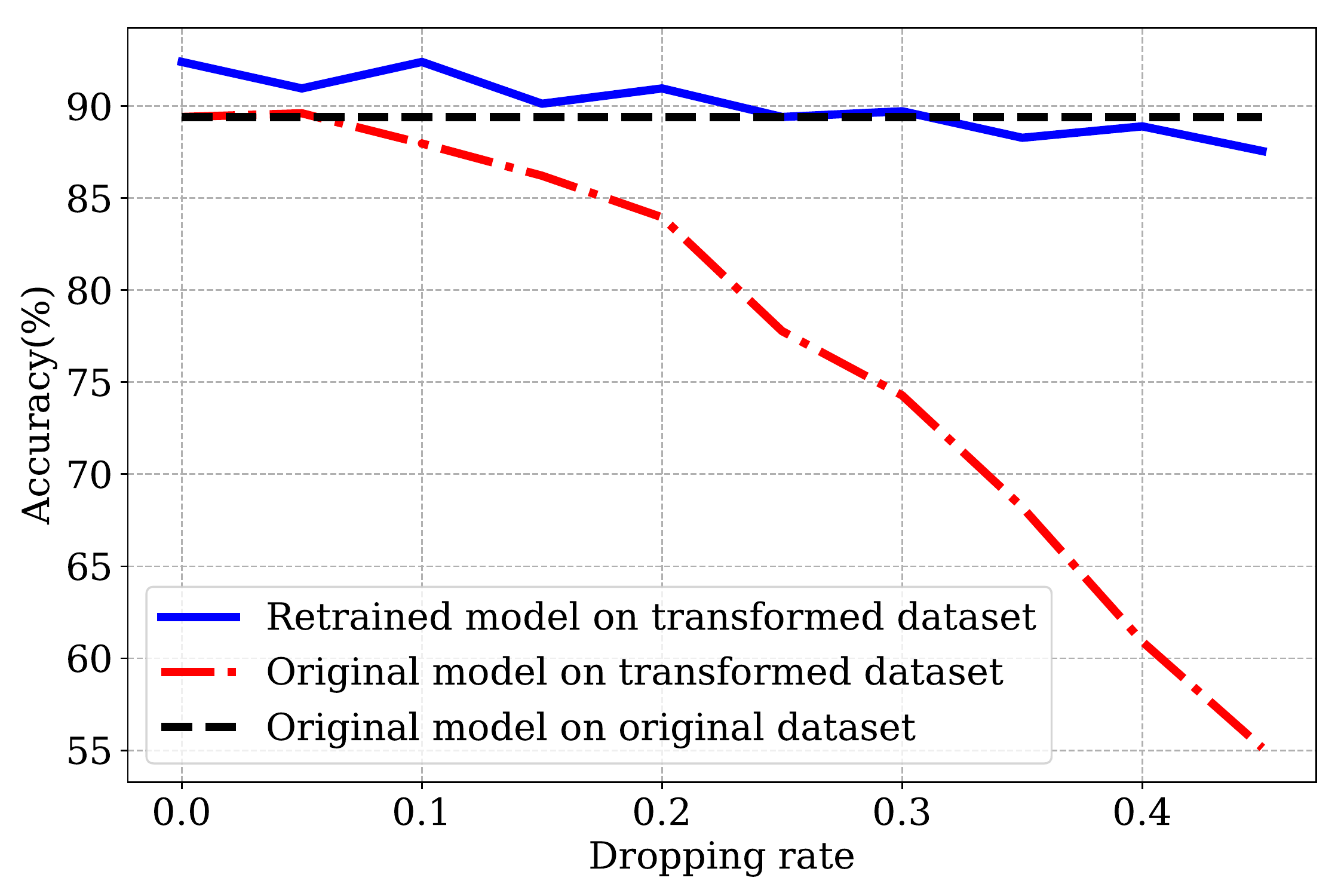}}
    \caption{
    The geometric domain shift is the key challenge of 3DDG. \textbf{Top:} Domain misalignment can be easily caused by broken assemblies or missing parts. \textbf{Bottom:} The existing deep networks learned in the complete point sets underperform in the transformed point sets with broken assemblies or missing parts. The X-axis represents the ratio of randomly discarded points for each point set.
    Further details of transformed datasets are included in Section \ref{sec:transform}.
    }
    \vspace{-15pt}
    \label{fig:front_figure}
\end{figure}

Understanding and reasoning about 3D objects are crucial for AI robots to understand the real world. Recently, with the success of deep learning in 2D vision tasks \cite{he2016deep,ren2015faster}, 3D deep learning techniques have also obtained state-of-the-art classification/detection results in both simulated and real-world datasets of point clouds \cite{qi2017pointnet,qi2017pointnet++,zhou2018voxelnet,qi2018frustum}.
However, the high performance of 3D deep learning is mostly achieved when the training data and testing data come from the same domain. But in practical applications, the assumption is always violated. 
For example, we can easily use the well-annotated CAD models collected from the Internet as training data \cite{wu20153d,chang2015shapenet}, but when we directly use the learned deep networks to classify real objects, we find that the performance decreases dramatically. 
Since scanning real objects using depth cameras or LiDAR sensors are extremely costly, it is worthwhile to study the Sim-to-Real transfer learning problem for 3D point clouds.

Under these circumstances, in this paper, we present a new problem of 3D Domain Generalization (3DDG), which aims to learn a classification model in source point sets that can achieve high performance in target point sets that are not accessible during training.
The training and testing data are from different domains with a large domain shift. In the case of 3DDG, it is mainly caused by various geometries of shapes from distant domains, namely the \textit{geometry shift}. 
Figure~\ref{fig:front_figure} (top) shows an example of geometry shift between the simulated ModelNet dataset \cite{wu20153d} and the real ScanObjectNN \cite{uy2019revisiting} dataset. 
There exist some misalignments of learned features caused by broken assemblies or missing parts of the entire point cloud.
Existing point cloud classification methods  \cite{ben20183dmfv,li2018pointcnn,wang2019dynamic,xu2018spidercnn,qi2017pointnet,qi2017pointnet++} perform sub-optimally under such a domain shift \cite{uy2019revisiting}.
In an early experiment of this paper, we empirically analyze their performance in target domains with incomplete point clouds.
The results of PointNet \cite{qi2017pointnet} are shown in Figure~\ref{fig:front_figure} (bottom).
Note that only by retraining it on the transformed target set with missing parts, can the model perform well on it. By contrast, if we directly evaluate a model trained on a full source dataset, the performance will degrade significantly, which can be caused by \emph{overfitting the geometry of training data}. 

To mitigate the geometric domain shift, we propose a meta-learning framework named MetaSets that has two contributions to 3DDG.
First and foremost, MetaSets improves the previous gradient-based meta-learning algorithms by applying a learned soft-sampling strategy to the meta-tasks. It is partly inspired by the idea that the robustness of the model to out-of-distribution test samples can be improved by penalizing pre-defined data groups with weights that are inversely proportional to the sample size of the groups \cite{sagawa2019distributionally}.
Specifically, we split the source dataset into a meta-training set and a meta-validation set, and perform meta-training and meta-validation on corresponding meta-tasks iteratively. 
In each meta-training phase, a batch of tasks is sampled with learned probabilities that are decided by the corresponding meta-validation error. 
This sampling strategy dynamically adjusts the training workload on all tasks and encourages the model to focus more on the difficult ones.
In this way, the proposed meta-learning algorithm can effectively balance the importance of different tasks.

Another contribution of MetaSets is the transformation approaches that are specifically designed for point cloud and can be used to build different meta-tasks by expanding the source dataset.
The transformed point sets simulate the cases of occlusions, missing parts, and the changes in scanning density in real environments. They can best facilitate the Sim-to-Real transfer learning by providing various geometric priors, and thus prevent the model from overfitting the source dataset.
In this way, the proposed algorithm can learn both domain-invariant latent factors that greatly improve the generalization ability of the model, and domain-specific geometric priors that enhance the discriminability.

We design two Sim-to-Real benchmarks based on standard point cloud datasets, and observe that the proposed MetaSets remarkably outperforms existing approaches for 3DDG. We further validate the effectiveness of the meta-learning framework, the soft-sampling strategy with learned probabilities, as well as each transformed point set.

\section{Problem Setup}

This work studies the problem of 3D Domain Generalization (3DDG), which has been missing a full study at present. 
In 3DDG, we have a source domain consisting of point clouds and labels $\mathcal{D}_\text{s}=\{(\mathbf{P}_\text{s}, y_\text{s})\}$, and the goal is to train a highly generalizable model $f$ to achieve low classification errors $\mathcal{E}=\mathbb{E}_{(\mathbf{P},y)\sim D_\text{t}}[f(\mathbf{P})\ne y]$ on an unseen target domain $\mathcal{D}_\text{t}=\{(\mathbf{P}_\text{t}, y_\text{t})\}$.
Only the data and labels in the source domain are available during training, while those in the target domain are only used for evaluation. The technical challenge of this problem is the variations of data distribution between $\mathcal{D}_\text{s}$ and $\mathcal{D}_\text{t}$, which usually exist in simulation-to-reality (Sim-to-Real) transfer learning scenarios due to geometry deformation or the change of the point cloud density, and may violate the i.i.d. assumption of the existing deep learning approaches for point cloud classification.
\section{MetaSets}

In this section, we first introduce the overall framework of a new meta-learning approach for 3D domain generalization named MetaSets. Then we describe the specific meta-tasks in the framework, from which the model can learn meta-level geometric priors and generalize the features to other unseen domains. 
Figure \ref{fig:arch} shows a schematic of our approach.

\subsection{The MetaSets Framework}
A possible solution to 3DDG is to mitigate the potential geometry shift of a variety of 3D point sets by learning the domain invariant geometric structures.
To this end, we propose the MetaSets framework, as shown in Alg. \ref{alg}, which follows the basic idea of Model-Agnostic Meta-Learning (MAML) \cite{finn2017model}, and contributes to ensuring the balance of various meta-tasks in the learning process so that features can be broadly generalizable to other unseen domains.
We use MAML because it is a typical gradient-based meta-learning approach, and MetaSets is supposed to be a general framework that can be combined with more advanced meta-learning algorithms.

\myparagraph{Meta-tasks.}

Since in practice we have only one source domain $\mathcal{D}_\text{s}$  of point sets that can be used for training, we need to augment $\mathcal{D}_\text{s}$ with reasonable data transformations to build multiple meta-tasks.
Assuming that there has been a set of transformation functions $\{F_n\}_{n=1}^N$ that we can use off-the-shelf, we may construct $n$ classification tasks $\{\mathcal{T}_n\}_{n=1}^N$ for the transformed point sets, and divide each of them into a meta-training set and a meta-validation set.
We will discuss the details of $F_n$ in Section \ref{sec:transform}.

\begin{algorithm*}[t]
\setstretch{1.15}
\caption{Training process of MetaSets}
\label{alg}
\begin{algorithmic}[1]
\Require Source dataset $\mathcal{D}_\text{s}=(\mathcal{D}_\text{s}^\text{train}, \mathcal{D}_\text{s}^\text{val})$, data minibatch size $B$, transformation functions $\{F_n\}_{n=1}^N$, number of sampled functions $K$, validation error bound $\epsilon$, learning rates $\eta$ and $\beta$
\State \textbf{Initialize} $\theta \leftarrow \theta_{0}$
\State $\{p_n\}_{n=1}^N=\frac{1}{N}$ \Comment{Initialize the task sample probability}
\While{$\mathcal{L}^\text{val}_{n}<\epsilon$ for $n=1,\ldots,N$}
\Repeat
\Comment{Meta-training phase}
\State $\{(\mathbf{P}_\text{s}^i,y_\text{s}^i)\}_{i=1}^B \sim \mathcal{D}_\text{s}^\text{train}$
\State $\{F_{k}^\prime\}_{k=1}^K \sim \mathrm{multinomial}(\{F_n\}_{n=1}^{N}; K) \  \text{w.r.t.} \ \{p_n\}_{n=1}^N$
\Comment{Sample $K$ functions}
\For{$k=1, \ldots, K$}
\State $\{\mathbf{P}_k^i\}_{i=1}^B = \{F_k^\prime(\mathbf{P}_\text{s}^i)\}_{i=1}^B$
\Comment{Obtain transformed point sets} %
\State $\mathcal{L}_{k} \leftarrow$ $\{f_\theta(\mathbf{P}_k^i)\}_{i=1}^B; \ \theta^\prime_{k} = \theta-\eta \nabla_\theta \mathcal{L}_{k}$ 
\Comment{According to
Eqn~\eqref{eqn:gradient}}
\EndFor
\State $\mathcal{L}^\text{cls} \leftarrow$  $\{f_{\theta_k^\prime}( \mathbf{P}_k^i )\}_{i=1,k=1}^{B,K}; \ \theta \leftarrow \theta - \beta \nabla_\theta \mathcal{L}^\text{cls}$ 
\Comment{According to Eqn~\eqref{eqn:meta-class}}
\Until{end of $\mathcal{D}_\text{s}^\text{train}$}
\Repeat
\Comment{Meta-validation phase}
\State $\{(\mathbf{P}_\text{s}^i,y_\text{s}^i)\}_{i=1}^B \sim \mathcal{D}_\text{s}^\text{val}$
\For{$n=1, \ldots, N$}
\State $\{\mathbf{P}_n^i\}_{i=1}^B = \{F_n(\mathbf{P}_\text{s}^i)\}_{i=1}^B$
\Comment{Obtain transformed point sets} %
\State $\mathcal{L}^\text{val}_n \leftarrow \{f_\theta(\mathbf{P}_n^i)\}_{i=1}^B$ 
\Comment{According to Eqn~\eqref{eqn:valid}}
\EndFor
\State $\{p_n\}_{n=1}^N \leftarrow \{\mathrm{softmax}(\{\mathcal{L}^\text{val}_n\}_{n=1}^N)\}_{n=1}^N$ 
\Comment{According to Eqn~\eqref{eqn:probability}}
\Until{end of $\mathcal{D}_\text{s}^\text{val}$}
\EndWhile
\State \textbf{return} $\ \theta$
\end{algorithmic}
\end{algorithm*}

\myparagraph{Meta-training.}

An inductive bias of MetaSets is that effective \textit{meta-learners} are supposed to perform well on all meta-tasks. 
For this purpose, in MetaSets, we improve the meta-training process of MAML \cite{finn2017model} by explicitly balancing the contributions of all meta-tasks to the meta-learner. 
As shown in \textbf{Line 6} in Alg. \ref{alg}, at each meta-training step, we dynamically sample $K$ transformation functions denoted by $\{F_k^\prime\}_{k=1}^K$ from the function set, instead of pre-progressing $\mathcal{D}_\text{s}$ in advance.
The probabilities of sampling $\{p_n\}_{n=1}^N$ are initialized with a uniform distribution across all the tasks and updated at each meta-validation step (\textbf{Line 17} in Alg. \ref{alg}). In other words, MetaSets not only learns to complete the predefined tasks but also learns to balance their impact on the meta-training process.
Based on a minibatch of point clouds $\{\mathbf{P}^i_\text{s}\}_{i=1}^{B}$ randomly sampled from the source meta-training set $\mathcal{D}_\text{s}^\text{train}$, MetaSets maps them to $K$ transformed point sets $\mathcal{D}_k^\text{train}=\{(\mathbf{P}^i_{k},{y}_\text{s}^i )\}_{i=1}^{B}$, where $k\in\{1,\ldots,K\}$ (\textbf{Line 8} in Alg. \ref{alg}). 
On each task, it computes the classification loss:
\begin{equation}\label{eqn:gradient}
    \mathcal{L}_{k} = \frac{1}{B}\sum_{i=1}^{B}\ell \left(f_\theta(\mathbf{P}^i_{k}), y^i_\text{s}\right),
\end{equation}
where $N$ is the minibatch size, $\ell(\cdot)$ is the cross-entropy loss, and $f_\theta$ is a point cloud classification model, \eg, PointNet \cite{qi2017pointnet}, parameterized by $\theta$. 
We compute the parameters after one gradient update as $\theta^\prime_{k}=\theta-\eta \nabla_\theta \mathcal{L}_k$, then use the following meta-objective to minimize the classification error over all sampled tasks with updated parameters $\theta^\prime_{k}$:
\begin{equation}\label{eqn:meta-class}
    \mathcal{L}^\text{cls} = \sum_{k=1}^{K}\frac{1}{B}\sum_{i=1}^{B}\ell \left(f_{\theta'_{k}}(\mathbf{P}^i_{k}), y^i_\text{s} \right).
\end{equation}

Minimizing the meta-objective achieves high performance on each sampled task, such that we can update the classification model by $\theta \leftarrow \theta - \beta \nabla_\theta \mathcal{L}^\text{cls}$.

\begin{figure*}[t]
    \centering
    \includegraphics[width=0.9\textwidth]{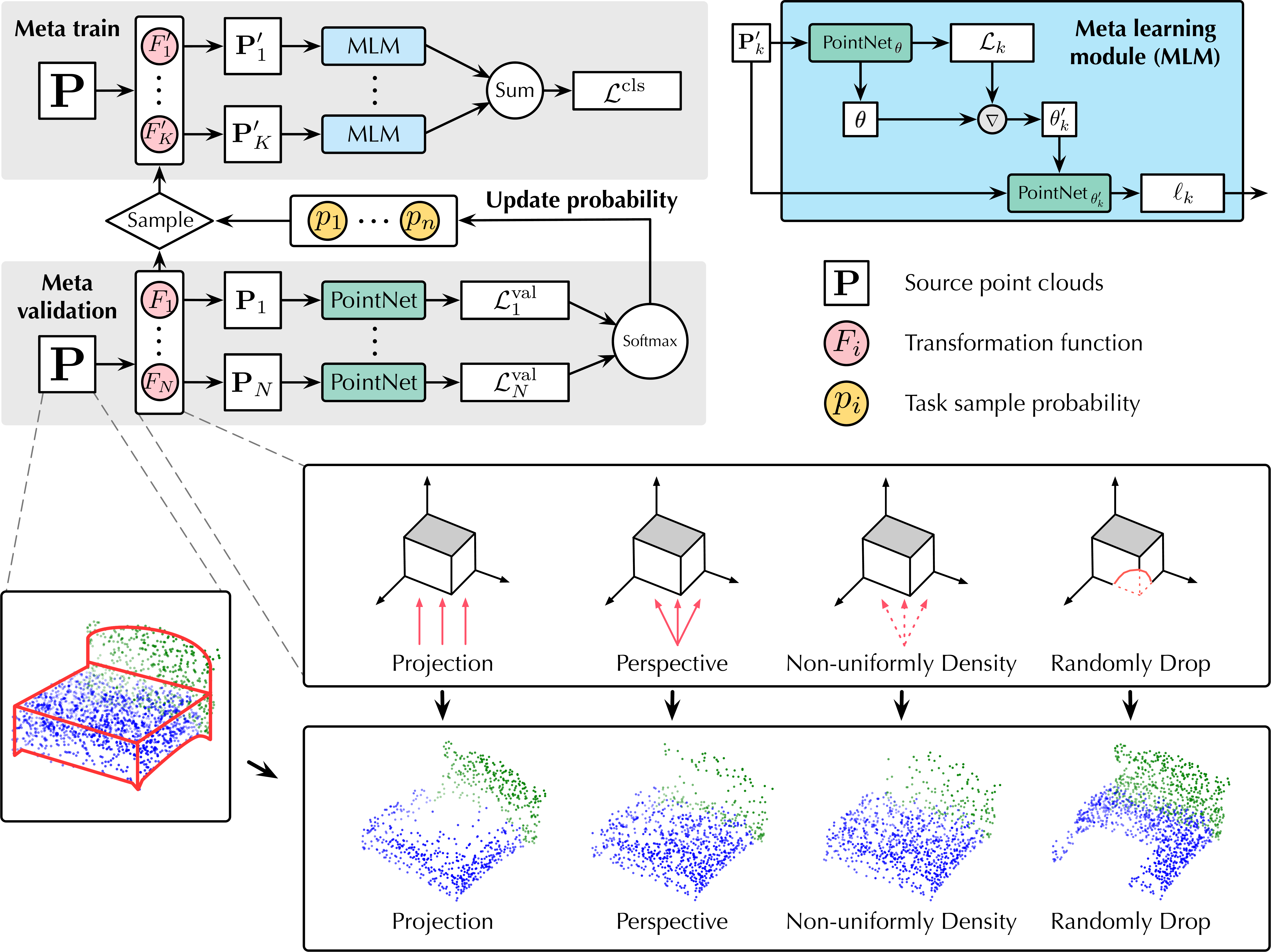}
    \vspace{3pt}
    \caption{A schematic of the meta-learning framework of MetaSets.}
    \label{fig:arch}
    \vspace{-5pt}
\end{figure*}

\myparagraph{Meta-validation.}

After one training epoch over $\mathcal{D}_\text{s}^\text{train}$, MetaSets performs the mata-validation process to evaluate the classification loss under parameter $\theta$ on $\mathcal{D}_\text{s}^\text{val}$ for the total of $N$ tasks. 
Concretely, we first sample $B$ point clouds from $\mathcal{D}_\text{s}^\text{val}$ and transform them into $\mathcal{D}_n^\text{val}$ for all $N$ transformation functions $\{F_n\}_{n=1}^N$, and then for each task $\mathcal{T}_n$, we have
\begin{equation}\label{eqn:valid}
    \mathcal{L}^\text{val}_{n}= \mathbb{E}_{(\mathbf{P}, y) \sim \mathcal{D}^\text{val}_{n}} \ \ell \big(f_\theta(\mathbf{P}), y \big).
\end{equation}

The validation loss $\mathcal{L}^\text{val}_{n}$ serves two purposes. First, it can be used as a signal of convergence for the classification model. If $\mathcal{L}^\text{val}_{n}$ is smaller than a small threshold $\epsilon$ (\ie, validation error bound) for all tasks $n\in \{1,\ldots,N\}$, the model is considered to be converged.

Second and more importantly, at the end of each meta-validation process, we use $\{\mathcal{L}^\text{val}_{n}\}_{n=1}^N$ to compute the probability of sampling each task for the next meta-training phase, as shown in Figure \ref{fig:arch}.
We propose to increase the probabilities of tasks with higher validation losses to be selected in the next meta-training process so that MetaSets can focus more on these challenging tasks.
Specifically, in \textbf{Line 17} in Alg. \ref{alg}, we update the probability of each task by performing softmax over $\{\mathcal{L}^\text{val}_{n}\}_{n=1}^N$:
\begin{equation}\label{eqn:probability}
    p_n = \frac{\exp(\mathcal{L}^\text{val}_{n})}{\sum_{n=1}^{N}\exp(\mathcal{L}^\text{val}_{n})}.
\end{equation}

The updated probabilities are then applied to the sampling process of the transformation functions and balance the importance of $N$ meta-tasks in the next training epoch. 
In 3DDG scenarios, learning $\{p_n\}_{n=1}^N$ dynamically prevents the classification model from overfitting the source domain, and encourages it to generalize across a wide range of possible variations of the source point clouds, such as geometric distortion and missing parts. 
Unlike the MAML algorithm \cite{finn2017model} that mainly focuses on encoding domain-invariant knowledge in the meta-leaner, the proposed MetaSets performs \textit{soft-sampling} on meta-tasks to further address the imbalance of their impact, and thus enables the model to learn both domain-invariant and domain-specific knowledge from the expanded source domains.

For the convergence of MetaSets, since the standard meta-learning paradigm has been proved to be able to converge theoretically \cite{fallah2020convergence} and empirically \cite{finn2017model} even for diverse tasks with a large number of data points and MetaSets follows a similar training paradigm to standard meta-learning, so MetaSets can also converge soundly. For the time cost, the proposed MetaSets only requires an extra meta-validation step compared to standard meta-learning. However, the meta-validation step costs much less time than the meta-training step (about $1:30$ on average in experiments). So the time cost of MetaSets is comparable to standard meta-learning. In experiments, we show that the convergence speed of MetaSets is similar to standard meta-learning and comparable to standard point cloud classification models.

\subsection{Transformed Point Sets}
\label{sec:transform}

The quality of the meta-tasks is the key to the effectiveness of MetaSets, as is the case in any other MAML variant.
To improve the generalization ability of the learned model, the tasks are supposed to cover a wide range of possible variations of point clouds from the source domain.
As mentioned above, we expand the source domain by applying a set of transformation functions $\{F_n\}_{n=1}^N$ to the original point clouds. 
At the same time, the diversity and representativeness of the transformed point sets are considered.
Due to the work of Wu \textit{et al.} \cite{wu2019squeezesegv2}, the common geometry shift for point clouds is mainly caused by occlusions, density changes, and scanning noises. 
Accordingly, we design the following three transformation approaches, as shown in Figure \ref{fig:arch}, that can be used to generate a variety of meta-tasks by changing the hyperparameters or using different combinations of them.

\myparagraph{Non-uniform density ($P_1,g$).} When scanning an object with a LiDAR, the closer it is, the denser the point cloud will be. 
We simulate the non-uniform density by firstly setting an anchor position $P_1$ outside the point cloud and calculating its distances to each point, where $P_1$ is randomly selected from a unit sphere. We then discard the points with probabilities (\ie, dropping rate) proportional to the distances. To construct different point sets, we first normalize the distance from $P_1$ to every point in the point cloud within the range of $[0,1]$, which is used as the basic drop rate. Then we multiply the basic drop rate with a multiplier $g>1$, which is called the gate parameter and controls the density of the point cloud, where a larger $g$ means more dropping points and thus a sparser point cloud. 
The point distribution and the density of the transformed point cloud can be controlled by $P_1$ and $g$ respectively.

\myparagraph{Dropping ($P_2,x\%$).} 
Inspired by the idea of dropout that keeps deep networks from overfitting, we propose to randomly drop parts of point cloud structures.
We randomly select a position $P_2$ in the point cloud and drop the nearest $x\%$ points, where $P_2$ and $x\%$ are controllable hyperparameters. 
This transformation method has two benefits. First, it produces meta-tasks that can prevent the model from overfitting the source domain. Second, it simulate the cases of broken assemblies or missing parts in real environments, and thus can best facilitate the Sim-to-Real generalization.

\myparagraph{Self-occlusion ($\vec{v},W$).} 
Self-occlusion occurs when the shape is viewed from a certain angle that the back surface of the point clouds is invisible. 
As shown in Figure \ref{fig:arch}, we simulate occlude by specifically designing a parallel projection method for point clouds, and take it as a basic operation to construct the meta-tasks. 
To be specific, we first choose a plain outside the 3D object and project the points cloud to the plain along its normal vector $\vec{v}$. 
All optional planes are supposed to have the same minimum distance to the point cloud.
We then divide the plane into grids of equal size. Within each grid, we keep the closest point along the normal vector and remove the others.
We can control the grid size of $W$ to obtain different occlusion patterns.
As it approaches infinity, it retains only one point, and when it is small enough, the entire point cloud will be preserved and no occlusion occurs.
We can also control the projection angle, \ie, the normal vector of the plane. 
The transformed point sets with self-occlusion can greatly benefit the transfer learning problem of 3D point clouds in Sim-to-Real scenarios as the inner structure of real-world data is not visible.

In summary, through the different types of transformations introduced above, we can simulate different geometry shifts of self-occlusions, density changes, and missing parts.
We control the hyperparameters in ($P_1,g,P_2,x\%,\vec{v},W$) to develop a set of functions $\{F_n\}_{n=1}^N$ at the very beginning of the training process.
Each of them can be used to construct a transformed point set $\mathcal{D}_n = \{(F_n(\mathbf{P}),y)\}$, where $(\mathbf{P},y)\in \mathcal{D}_\text{s}$.
However, it is worth noting that, the imbalance of the task difficulty may make the meta-learning approach less effective for 3DDG.
To solve this problem, we introduce the \textit{soft-sampling} technique to dynamically adjust the training workload of the model on each task.

Another benefit of the above transformation methods is that we design tasks at the input level so that they can be easily combined with existing approaches that have specific designs in model architectures \cite{qin2019pointdan,rao2020global}.
Besides, the overall MetaSets framework is extendable and can be generalized to other target domains by simply defining new tasks with specific geometric priors.

\section{Experiments}

To evaluate MetaSets in Sim-to-Real scenarios, we build two 3DDG benchmarks upon the synthetic datasets, ModelNet \cite{wu20153d} and ShapeNet \cite{chang2015shapenet}, as well as a real dataset ScanObjectNN \cite{uy2019revisiting}. 
See \underline{supplementary materials} for more details.

\myparagraph{Compared methods.}
We compare MetaSets with five state-of-the-art point cloud classification models: PointNet~\cite{qi2017pointnet}, PointNet++~\cite{qi2017pointnet++}, DGCNN \cite{wang2019dynamic}, ConvPoint~\cite{BOULCH202024}, LDGCNN~\cite{zhang2019linked}, and PointCNN~\cite{li2018pointcnn}, covering both PointNet-based, CNN-based and Graph-based methods. 
Besides, we further compare MetaSets with PointDAN \cite{qin2019pointdan}, a domain adaptation approach for point clouds based on PointNet.
It is worth mentioning that PointDAN \cite{qin2019pointdan} requires unlabeled target data for training, while for 3DDG, the distribution of target data is not accessible. Thus, compared with PointDAN \cite{qin2019pointdan}, MetaSets tries to tackle a more challenging problem. 

\myparagraph{Implementation details.}
We divide each source domain into a training set and a validation set at a scale of $5:1$, following the experimental protocol of domain generalization \cite{ghifary2015domain}. Similar to \cite{finn2017model}, we implement our meta-learning algorithm with first-order approximation to achieve high computational efficiency. For all the compared methods, we tune hyperparameters by performing cross-validation in the source domain. 
For MetaSets, we construct three tasks for each transformation method, where the parameters are selected as follows: (1) we monotonically change the parameters, from preserving all points to deleting all points; When we observe that the shape of the resulting point cloud begins to change, we record the parameters as $t_1$, and when we observe that the shape is almost unrecognizable, we record the parameters as $t_2$. (2) We randomly sample three parameters within the range of $(t_1,t_2)$ and add them to the task set.
At each iteration in the meta-training phase, we sample four tasks from the task set of all transformations. Hyperparameters $\eta$ and $\beta$ are tuned by cross-validation, as shown in Alg. \ref{alg}. Please see \underline{supplementary materials} for their sensitivity analysis. 
Unless otherwise specified, we use PointNet \cite{qi2017pointnet} as the backbone of MetaSets. 
All experiments are conducted using PyTorch \cite{NEURIPS2019_9015}. We perform each experiment three times and report the average and the standard deviation of the results.

\subsection{Geometry Overfitting}

We first empirically study how our current point cloud classification methods are affected by geometry shift.
As shown in Figure \ref{fig:front_figure} (please go back to the first page), we evaluate PointNet~\cite{qi2017pointnet} on the ModelNet dataset by randomly dropping some parts of the point cloud. 
As we can see, the classification accuracy becomes lower as more points are deleted.
One might argue that this performance degradation is because the remainder of the point cloud does not have sufficient geometric information and is, therefore, less recognizable.
Thus, we re-train the PointNet on transformed point clouds and the performance rises back to the model on the original dataset. 
The results show that the point clouds after transformation are still recognizable, and the reason for the performance degradation is the geometry shift between training and testing, which is a common situation in most Sim-to-Real scenarios.
The model overfits the geometry of the complete point cloud shapes and cannot recognize the geometry of transformed shapes.

\begin{figure*}[t]
    \centering
    \includegraphics[width=\textwidth]{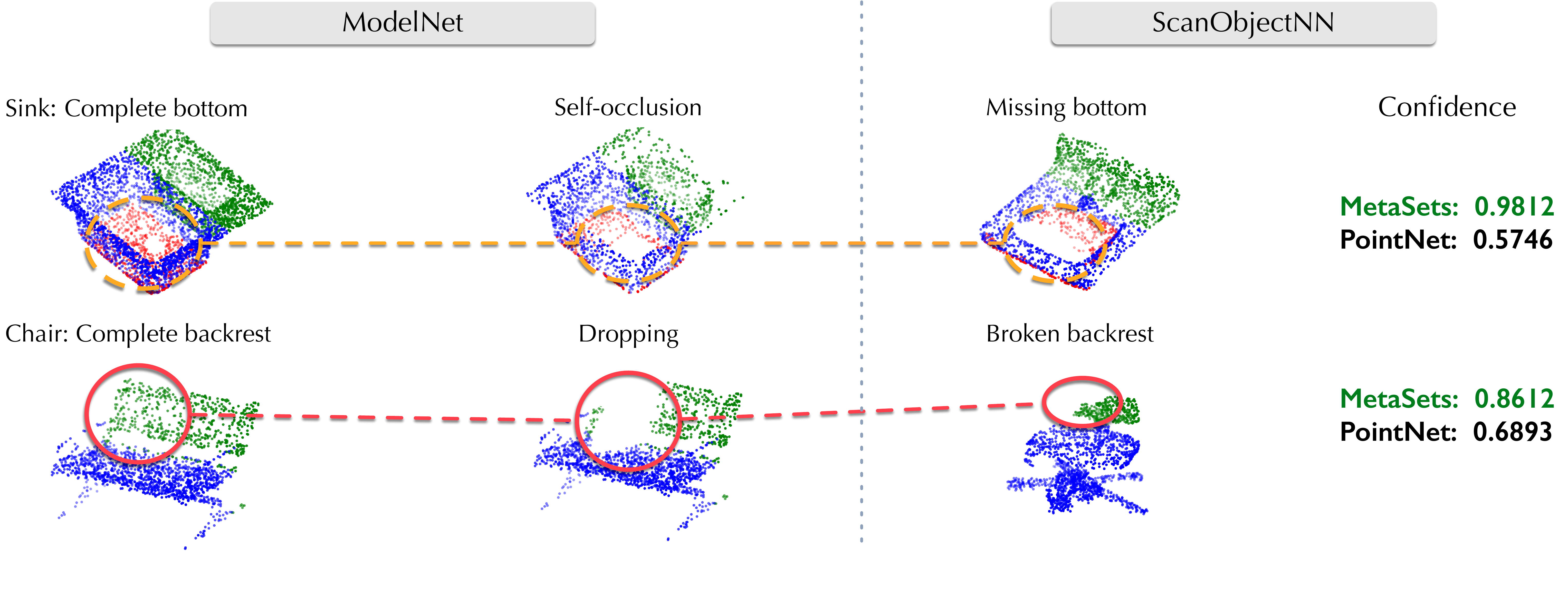}
    \vspace{-30pt}
    \caption{Visualization of point clouds from the source domain (\textbf{first column}), the transformed point sets (\textbf{second column}), and the target domain (\textbf{third column}). The geometries of the transformed point clouds match well with those of the target clouds, which enables the proposed MetaSets to alleviate the influence of geometric domain shift.}
    \label{fig:visual}
    \vspace{-5pt}
\end{figure*}

\subsection{ModelNet to ScanObjectNN}

On this benchmark, we select the $11$ categories shared by the ModelNet40 and ScanObjectNN datasets. Models are trained on ModelNet and evaluated on ScanObjectNN. We report the class-wise results in the supplementary materials.

\myparagraph{Visualization of transformed point sets.} 

Figure \ref{fig:visual} illustrates the differences between geometries in the synthetic source domain and those in the real target domain, and how the transformed point sets contribute to learning more generalizable features.
Due to self-occlusion or missing parts, these point clouds that are randomly sampled from the real domain suffer from ``missing bottom'' and ``broken backrest'', which leads to the decrease of the classification confidence of the original PointNet. 
In contrast, the transformed point sets proposed in this work can help bridge the geometry gap between complete and imperfect point clouds, enabling the meta-learning model to learn representations that can be easily generalized to the geometries in the target domain.

\begin{table}[t]
    \centering 
    \resizebox{\linewidth}{!}{
    \begin{tabular}{|l|cc|}
    \hline
    \multirow{2}{40pt}{Method} & \multirow{2}{40pt}{Object} & Object \& \\
     &  &  Background \\
    \hline\hline
    PointNet~\cite{qi2017pointnet} & 55.90$\pm$1.47 & 49.48$\pm$2.28\\
    PointNet++~\cite{qi2017pointnet++} & 47.30$\pm$0.53 & 40.42$\pm$1.17\\
    ConvPoint~\cite{BOULCH202024} & 57.40$\pm$0.44 & 55.44$\pm$0.32 \\
    DGCNN~\cite{wang2019dynamic} & 61.68$\pm$1.26 & 57.61$\pm$0.44 \\
    PointCNN~\cite{li2018pointcnn} & 50.32$\pm$0.43 & 46.11$\pm$0.43 \\
    LDGCNN~\cite{zhang2019linked} & 62.29$\pm$0.22 & 58.83$\pm$0.43\\
    \hline
    PointDAN~\cite{qin2019pointdan} on PointNet & 63.32$\pm$0.85 & 55.13$\pm$0.97\\
    \hline
    MetaSets on PointNet & 68.28$\pm$0.79 & 57.19$\pm$1.23 \\ 
    MetaSets on ConvPoint & 65.05$\pm$0.56 & 61.33$\pm$0.32 \\
    MetaSets on DGCNN & \textbf{72.42}$\pm$0.21 & \textbf{65.66}$\pm$1.06 \\
    \hline
    \end{tabular}
}
\vspace{2pt}
\caption{Accuracy (\%) on the ModelNet$\rightarrow$ScanObjNN benchmark.}
    \label{tab:mn_sonn}
    \vspace{-5pt}
\end{table}

\begin{table*}[t]
    \centering 
    \renewcommand\tabcolsep{5pt}
    \small
    \begin{tabular}{|l|ccccc|}
    \hline
    Method & None & Density & Dropping & Self-occlusion & All tasks \\
    \hline\hline
    MetaSets on PointNet & \textbf{58.95}  & \textbf{65.47} & \textbf{64.00} & \textbf{63.58} & \textbf{68.28} \\
    MetaSets w/o meta-learning & 55.90 & 62.58 & 62.95 & 61.47 & 63.16  \\
    \hline
    \end{tabular}
    \vspace{5pt}
    \caption{The ablation study of using different meta-tasks (\textbf{columns}) and different optimization algorithms (\textbf{rows}) on ModelNet$\rightarrow$ScanObjNN.
    }
    \label{tab:ablation}
\end{table*}

\begin{table*}[t]
    \centering 
    \small
    \renewcommand\tabcolsep{10pt}
    \begin{tabular}{|l|cc|cc|cc|cc|}
    \hline
    \multirow{2}{100pt}{Meta-training method} & \multicolumn{2}{c|}{Density} & \multicolumn{2}{c|}{Dropping} & \multicolumn{2}{c|}{Self-occlusion} & \multicolumn{2}{c|}{Avg} \\
    & Acc (\%) & Loss & Acc (\%)& Loss & Acc (\%)& Loss & Acc (\%)& Loss \\
    \hline \hline
    Density & 59.58 & 1.49 & 57.26 & 1.61 & 62.53 & 1.42 & 59.79 & 1.51 \\
    Dropping  & 56.63 & 1.68 & 58.30 & 1.69 & 62.31 & 1.55 & 59.08 & 1.64 \\
    Self-occlusion  & 51.16 & 1.85 & 53.73 & 1.93 & 62.74 & 1.68 & 55.88 & 1.82 \\
    \hline
    \textbf{Final:} MetaSets & \textbf{68.42} & \textbf{1.13} & \textbf{65.05} & \textbf{1.20} & \textbf{63.16} & \textbf{1.26} & \textbf{65.54} & \textbf{1.20} \\
    MetaSets w/o soft-sampling & 67.09 & 1.35 & 60.07 & 1.51 & 60.35 & 1.49 & 62.50 & 1.45 \\
    \hline
    \end{tabular}
    \vspace{5pt}
    \caption{Generalization analysis across meta-tasks on the ModelNet$\rightarrow$ScanObjNN benchmark. 
    See text for more details.
    }
    \label{tab:generalization_accuracy}
    \vspace{-5pt}
\end{table*}

\myparagraph{Comparison with the single-domain state of the art.}

As we can see from Table \ref{tab:mn_sonn}, MetaSets outperforms previous point cloud classification methods, including the state-of-the-art LDGCNN, under the 3DDG setting. 
The sub-optimal performance of previous single-domain approaches, which are also trained on the source domain only, is caused by overfitting the geometry of complete point clouds in the training set and thus lacking the ability to generalize to the real domains with imperfect shapes.
MetaSets obtains remarkable and consistent improvement, indicating that it is model-agnostic and performs better when the backbone network gets stronger.

\myparagraph{Comparison with PointDAN \cite{qin2019pointdan}.}

PointDAN is a domain adaptation model that incorporates target unlabeled data into training. 
However, as shown in Table \ref{tab:mn_sonn}, it underperforms MetaSets by $63.32\%$ vs. $68.28\%$.
Note that (1) MetaSets and PointDAN use the same backbone: PointNet, and (2) MetaSets does not require any target data throughout the training. 
Thanks to the geometry priors learned from various ``imagined'' point clouds from the source domain, MetaSets tries to solve a more challenging problem while turns out to achieve superior results. Therefore, we can consider MetaSets as a good solution to the 3DDG problem.

\myparagraph{Ablation study on meta-tasks.} 

Table \ref{tab:ablation} compares the effectiveness of each individual meta-task. We respectively train the model on the source dataset without any transformed data (the ``None'' column), the expanded source dataset with data transformation, and the one with all types of transformed data (the ``All tasks'' column). 
The first row is the proposed MetaSets that integrates geometrical transformations with the meta-learning approach based on soft-sampling. 
The second row, ``w/o meta-learning'', means learning directly on the above training dataset rather than using any meta-learning algorithms. 
It is worth noting that the soft-sampling technique is only effective if the number of tasks in the meta-learning process is greater than $2$.
Particularly, for the ``None'' column, the difference between the two models is that MetaSets is trained with the MAML algorithm on the raw dataset while the ``w/o meta-learning'' baseline model is trained using the na\"ive SGD, as opposed to bilevel optimization (L9-10 in Alg.~\ref{alg}).
For both compared models, we can observe that each individual transformation method (as the meta-task or data augmentation) independently brings in performance gains upon the ``None'' column, and the best performance is achieved when using all of them, indicating that these transformation methods are complementary to each other for inferring different geometry priors.

\myparagraph{Generalization across meta-tasks.} 

We further evaluate the generalization ability of MetaSets even across different transformed sub-datasets. For each of the first three rows in Table \ref{tab:generalization_accuracy}, we only use one type of transformation method to build the meta-task, and then evaluate the model on point sets produced by different types of transformation methods. 
For the last two rows, we use the full set of meta-tasks in the training process.
Each column indicates the task on which meta-validation is performed, where we report the meta-validation accuracy (\textit{Acc}) and cross-entropy loss (\textit{Loss}).
The results show that the model learned from the point clouds after non-uniform density transformation can be better generalized to other types of data with different geometry variations. The final results are further improved by using all three transformation techniques to build meta-tasks.

\myparagraph{Ablation study on model components.}

From Table \ref{tab:ablation}, MetaSets trained with the proposed meta-learning algorithm significantly outperforms the ``w/o meta-learning'' baseline model that uses the transformed point sets to augment the source dataset. The results validate the effectiveness of meta-learning.
Further, Table \ref{tab:generalization_accuracy} compares the final MetaSets with a baseline model that is trained with all meta-tasks but without soft-sampling.
It can be observed that soft-sampling reduces the cross-entropy loss of meta-validation remarkably. In fact, the classification difficulty is not the same on all transformed point sets, and the proposed soft-sampling method allows the training process to focus more on the difficult ones, so as to reduce the meta-validation loss of all tasks more effectively.

\subsection{ShapeNet to ScanObjetNN}

To ensure that the performance gain is not due to the bias of our method to a particular benchmark, we conduct experiments on ShapeNet to ScanObjectNN. In this benchmark, we train on the ShapeNet and evaluate on the ScanObjectNN. We select the $9$ categories shared by ShapeNet and ScanObjectNN.
As shown in Table \ref{tab:sn_sonn}, our MetaSets still outperforms all the other methods testing on objects either without or with a background. In particular, MetaSets achieves a larger margin on objects with background, which has a large domain shift from the source domain. Note that with a more advanced backbone (DGCNN vs. PointNet), the margin between MetaSets and the backbone is larger, which indicates an advanced backbone better utilizes the features from MetaSets and MetaSets has the potential to achieve higher performance with a better backbone in the future.

\begin{table}[t]
    \centering 
    \small
    \begin{tabular}{|l|cc|}
    \hline
    \multirow{2}{40pt}{Method} & \multirow{2}{40pt}{Object} & Object \& \\
     &  &  Background \\
    \hline \hline 
    PointNet~\cite{qi2017pointnet} & 54.00$\pm$0.32 & 45.50$\pm$0.99 \\
    PointNet++~\cite{qi2017pointnet++} & 45.50$\pm$0.64 & 43.25$\pm$1.23 \\
    ConvPoint~\cite{BOULCH202024} & 52.58$\pm$0.58 & 50.67$\pm$0.88 \\
    DGCNN~\cite{wang2019dynamic} & 57.42$\pm$1.01 & 54.42$\pm$0.80 \\
    PointCNN~\cite{li2018pointcnn} & 49.42$\pm$0.29 & 43.92$\pm$0.63 \\
    LDGCNN~\cite{zhang2019linked} & 57.92$\pm$0.63 & 52.50$\pm$0.25\\
    \hline
    PointDAN~\cite{qin2019pointdan} on PointNet & 54.95$\pm$0.87 & 43.00$\pm$0.95 \\
    \hline
    MetaSets on PointNet & 55.25$\pm$0.35 & 49.50$\pm$0.43 \\ 
    MetaSets on DGCNN & \textbf{60.92}$\pm$0.76 & \textbf{59.08}$\pm$1.01 \\
    \hline
    \end{tabular}
    \vspace{5pt}
    \caption{Accuracy (\%) on the ShapeNet$\rightarrow$ScanObjNN benchmark.
    }
    \label{tab:sn_sonn}
    \vspace{-10pt}
\end{table}

\section{Related Work}

\paragraph{Point cloud classification.}
Deep learning methods for 3D shape data were originally designed for 3D meshes \cite{wu20153d}, multi-view images \cite{su2015multi,qi2016volumetric,cao20173d}, and voxels \cite{maturana2015voxnet}. 
More recently, the emergence of point clouds has encouraged a growing number of deep networks specifically designed for this simple 3D data structure.
PointNet addresses the permutation invariance of point clouds by leveraging the max pooling \cite{qi2017pointnet}.
PointNet++ improves PointNet by encoding point clouds hierarchically at multiple scales \cite{qi2017pointnet++}.
Another focus of recent approaches is to develop the convolution operation applied to point clouds \cite{li2018pointcnn,wang2019dynamic,xu2018spidercnn,wu2019pointconv}. 
However, all of these methods assume that both the training and testing data are from the same domain.
When generalized across domains, these methods tend to overfit the geometry of the source domain and significantly degrade performance in the target domain.
Therefore, a new method is needed to address the domain generalization problem for 3D point clouds.

\myparagraph{Learning generalizable features.}
The domain generalization (DG) problem is to train a model that can be generalized to unseen target domains, which has been studied extensively in the field of image classification \cite{muandet2013domain}.
Most existing methods focus on learning domain invariant representations \cite{muandet2013domain,ghifary2015domain,li2018domain,li2018learning,li2019episodic}.
Khosla \textit{et al.} \cite{khosla2012undoing} and Li \textit{et al.} \cite{li2017deeper} used a hierarchy of parameters composed of domain-agnostic and domain-specific parts.
Shankar \textit{et al.} \cite{shankar2018generalizing} and Volpi \textit{et al.} \cite{volpi2018generalizing} augmented the training domain with adversarial perturbations. 
Li \textit{et al.} \cite{li2017learning} proposed to split source domains into meta-train and meta-test splits and conduct standard MAML across splits without data transformations.
However, all of these methods are only validated on image datasets but may fail on the 3D data due to the large gap between 2D images and 3D point clouds.
Previous work for image classification has shown that some heavy data augmentation techniques, such as color jittering and rotation, in some cases can be more helpful than gradient-based adversarial perturbations, when the goal is to improve domain generalization performance \cite{volpi2019addressing}. Similarly, the jittering method has also been used for point clouds by PointNet \cite{qi2017pointnet} for the general data augmentation purpose, but this approach does not explicitly consider any possible data discrepancy across domains. Different from the prior work, we propose three transformation approaches which are complementary to each other and particularly designed for 3DDG. These approaches show better performance than straightforward 3D data augmentation.

\myparagraph{Cross-domain point cloud classification.} 

As the deep networks achieve remarkable progress in recognizing synthetic point clouds, the research focus in this field is gradually shifting to real-world data by improving the transferability of deep models.  
But, as shown in \cite{uy2019revisiting}, the performance degrades due to the existence of a simulation-to-reality gap, such as geometry deformation or the change of the point cloud density.
In order to reduce the domain gap, PointDAN minimizes the Maximum Mean Discrepancy across domains \cite{qin2019pointdan}. However, it can only be used as a solution to the problem of domain adaptation with known target domain data distribution.
Unlike the above methods, our work provides an early study of 3DDG by performing meta-learning over the \textit{transformed point sets} that contain a variety of geometry priors of point clouds for generalizable representations.

\section{Conclusion}
This paper presented a new research problem of 3DDG, which aims to learn generalizable models that can be transferred to unseen target domains.
Due to the geometry shift, the previous deep models tend to overfit the geometry of the synthetic point clouds and deteriorate on the target real-world data. 
In this work, we proposed MetaSets.
First, it improves MAML by introducing a soft-sampling technique for the meta-tasks, which dynamically adjusts the training workload on each task, and enables the model to focus on the more difficult tasks. 
Second, we defined a set of basic data transformation methods to construct the meta-tasks, such that the transformed point sets can cover various point cloud geometries in real domains.
We designed two Sim-to-Real 3DDG benchmarks, and demonstrated that MetaSets remarkably outperformed the state-of-the-art point clouds classification models, even including a domain adaptation approach that uses target data for training.

\section*{Acknowledgments}

This work was fully supported by the AI project granted by China's Ministry of Industry and Information. Yunbo Wang was supported in part by CAAI-Huawei MindSpore Open Fund.

\appendix

\section{Experimental Details}
We introduce additional experiment details including dataset details and hyper-parameters details in this section.
\subsection{Dataset Construction}
\subsubsection{Sim-to-Real dataset}
Our Sim-to-Real dataset consists of three domains: ModelNet, ShapeNet and ScanObjectNN.

\noindent\textbf{ModelNet}\footnote{https://modelnet.cs.princeton.edu/}  
is a comprehensive clean collection of 3D CAD models of $40$ common object categories in the world. The CAD models are collected from online search engines by querying for each object category term. The dataset has two forms: ModelNet40 and ModelNet10. ModelNet40 contains all the CAD models of all the categories while ModelNet10 contains $10$ popular object categories from the $40$ categories, where models that did not belong to these categories are manually deleted. We use ModelNet40 in the experiments, and use the official training set as the meta-training set.  

\noindent\textbf{ShapeNet}\footnote{https://www.shapenet.org/}  
consists of several subsets: ShapeNetCore for classification and ShapeNetSem for segmentation. We use ShapeNetCore in the experiments. ShapeNetCore is a subset of the full ShapeNet dataset with single clean 3D models and manually verified category and alignment annotations. It covers $55$ common object categories with about $51{,}300$ unique 3D models. We use the official split of training and validation as our training and validation set.

\noindent\textbf{ScanObjectNN}\footnote{https://hkust-vgd.github.io/scanobjectnn/} 
is a new real-world point cloud object dataset based on scanned indoor scene data, which is built on two popular scene meshes datasets: SceneNN \cite{hua2016scenenn} and ScanNet \cite{dai2017scannet}. It contains about $15{,}000$ objects of $15$ categories with $2{,}902$ unique object instances. We use the official split with about $80\%$ training shapes and about $20\%$ testing shapes.

For both ModelNet40 and ShapeNet, we adopt the method from Qi \textit{et al.} \cite{qi2017pointnet} to generate the point clouds.
The shared categories in each dataset are shown in Table~\ref{tab:share_category}. As for data pre-processing, we follow the work from Qin \textit{et al.} \cite{qin2019pointdan} to normalize the point clouds into a unit ball.

\begin{table}[t]
    \centering 
    \caption{Selected categories for each domain generalization benchmark.}
    \label{tab:share_category}
    \vspace{5pt}
    \renewcommand\tabcolsep{10pt}
    \begin{tabular}{cc}
    \toprule
     ModelNet$\rightarrow$ScanObjNN & ShapeNet$\rightarrow$ScanObjNN \\
    \midrule 
    Bed & Bag \\
    Cabinet (Dresser, Wardrobe) & Bed \\
    Chair (Bench, Chair, Stool) & Cabinet \\
    Desk & Chair \\
    Display (Monitor) & Display \\
    Door & Pillow \\
    Shelf (Bookshelf) & Shelf (Bookshelf) \\
    Sink & Sofa\\
    Sofa & Table\\
    Table & - \\
    Toilet & - \\
    \bottomrule
    \end{tabular}
\end{table}

\begin{table*}[t]
    \centering 
    \small
    \renewcommand\tabcolsep{6pt}
        \begin{tabular}{|l|cccccccccccc|}
        \hline
        Method & bed & cabinet & chair & desk & display & door & shelf & sink & sofa & table & toilet & Avg\\
        \hline\hline
        PointNet~\cite{qi2017pointnet} & 40.91 & 1.33 & \textbf{97.44} & 50.00 & 61.90 & \textbf{100.00} & 61.22 & 37.50 & 52.38 & 51.85 & 29.41 & 53.09 \\
        MetaSets on PointNet &\textbf{63.64} & \textbf{18.67} & 94.87 & \textbf{53.33} & \textbf{69.05} & 95.24 & \textbf{75.51} & 37.50 & \textbf{71.43} & \textbf{74.07} & \textbf{64.71} & \textbf{65.27}\\
    \hline
    \end{tabular}
    \vspace{5pt}
    \caption{Accuracy per class (\%) and the average on ModelNet$\rightarrow$ScanObjNN. 
    }
    \label{tab:class_wise}
\end{table*}
\subsection{Hyperparameters for Transformed Point Sets}

As stated in the main text, we first confirm a valid hyperparameter range that changes the geometry of the shape but still make it recognizable, and then we randomly sample in the range. We try different random sampling method. We find that total random sampling sometimes samples similar hyperparameters, which decreases the diversity of the geometry priors contained in each transformed dataset. Therefore, we use a stratified sampling strategy that first evenly divides the valid hyperparameter range into $3$ sub-ranges, where $3$ is the number of specific transformations for each type of transformation. Then we randomly sample a hyperparameter in each sub-range.
We use the same transformation hyperparameters for ModelNet$\rightarrow$ScanObjNN and ShapeNet$\rightarrow$ScanObjNN, which are shown in Table~\ref{tab:parameter_trans}.

\begin{table}[h]
    \centering 
    \caption{Hyperparameters for different transformed point sets.}\label{tab:parameter_trans}
    \vspace{5pt}
    \renewcommand\tabcolsep{4pt}
    \begin{tabular}{lccc}
    \toprule
    Transformation & Param 1 & Param 2 & Param 3 \\
    \midrule 
    Self-occlusion (grid size $W$) & 0.035 & 0.022 & 0.017 \\
    Non-uniform density (gate $g$) & 1.3 & 1.4 & 1.6 \\
    Dropping (drop ratio $x\%$) & 24\% & 36\% & 45\% \\
    \bottomrule
    \end{tabular}
\end{table}

\subsection{Training Hyperparameters}

In all experiments, the validation convergence bound $\epsilon$ is set to $0.1\%$, and the batch size is set to $128$. We use Adam optimizer \cite{kingma2014adam}.
For all benchmarks, $\eta$ is $0.0003$, $\beta$ is $0.001$.
We conduct experiments on a machine with $64$ CPUs and $4$ GeForce 2080ti GPUs. The total training of $18{,}000$ iterations uses about $16$ hours.

\section{More Experimental Results}
In this section, we show more experiment results including class-wise results in the ModelNet$\rightarrow$ScanObjNN dataset, variants of soft-sampling, variants of meta-training and meta-objective, static or dynamic transformation and hyper-parameter sensitivity.

\subsection{Class-wise results.}
Table~\ref{tab:class_wise} shows the class-wise classification accuracy. We can observe that MetaSets outperforms the backbone network PointNet remarkably on most classes.
In particular, on difficult classes such as \textit{cabinet}, PointNet yields extremely low accuracy due to the large domain shift, \eg, the cabinets from the synthetic dataset has complex and detailed inner structures that are probably invisible in the real dataset. 
However, the expanded tasks of MetaSets can induce a larger set of geometry priors, which have a higher chance to include geometry priors that are similar to those from the target domain. Such priors enable MetaSets to mitigate the domain shift and perform still strongly for difficult cases of imperfect point clouds.
Even though on some classes like chair and door, MetaSets does not perform as well as PointNet. We manually check the objects of these classes in ModelNet and ScanObjectNN datasets and find that the objects are in different shapes like different kinds of chairs. So the classes suffer from huge domain shift but not geometric shift, which can only be addressed by access to the target data and is not the focus of the paper.

\begin{table}[ht]
    \centering 
    \caption{Variants of Meta-Validation and Meta-Objective. 'Per-batch $\mathcal{D}^\text{train}_\text{s}$' means meta-validation on a batch of meta-training set. '$\mathcal{D}^\text{train}_\text{s}$' means meta-validation on the meta-training set $\mathcal{D}^\text{train}_\text{s}$. 'Per-batch $\mathcal{D}^\text{val}_\text{s}$' means meta-validation on a batch of meta-validation set. '$\mathcal{D}^\text{train}_\text{s}$' means meta-validation on the meta-validation set $\mathcal{D}^\text{val}_\text{s}$, which is the proposed MetaSets.}\label{tab:soft-sampling}
    \renewcommand\tabcolsep{7pt}
    \small
    \begin{tabular}{cccc}
    \toprule
     Per-batch $\mathcal{D}^\text{train}_\text{s}$ & $\mathcal{D}^\text{train}_\text{s}$ & Per-batch $\mathcal{D}^\text{val}_\text{s}$ & $\mathcal{D}^\text{val}_\text{s}$ (Proposed)  \\
    \midrule
    63.16 & 64.42 & 66.53 & 68.28 \\
    \bottomrule
    \end{tabular}
\end{table}

\subsection{Variants of Soft-sampling}
We further explore the soft-sampling algorithm. We compare performance of soft-sampling probabilities computed by meta-validation based on different sets: a mini-batch of $\mathcal{D}^\text{train}_\text{s}$, the whole $\mathcal{D}^\text{train}_\text{s}$, a mini-batch $\mathcal{D}^\text{val}_\text{s}$, and the whole $\mathcal{D}^\text{val}_\text{s}$ (the proposed MetaSets). As shown in Table~\ref{tab:soft-sampling}, computing the soft-sampling probabilities based on the whole $\mathcal{D}^\text{val}_\text{s}$ outperforms all the other variants. This can be explained by that computing the probability of a batch of data is unstable while meta-validation on $\mathcal{D}^\text{train}_\text{s}$ causes the overfitting problem. 

\begin{table}[ht]
    \centering 
    \caption{Variants of Meta-Training and Meta-Objective. 'Mixture' means mixing the data of all three transformations into a single task and conduct meta-learning on the single task. 'Maximum Loss' means minimizing the maximum meta-training loss among all tasks.}\label{tab:meta-train}
    \renewcommand\tabcolsep{7pt}
    \small
    \begin{tabular}{ccc}
    \toprule
     Mixture & Maximum Loss & MetaSets  \\
    \midrule
    63.47 & 61.89 & 68.28 \\
    \bottomrule
    \end{tabular}
\end{table}

\subsection{Variants of Meta-Training and Meta-Objective}
One close variant of the proposed MetaSets is conduct meta-learning on the mixture of the data from three transformations, which uses no soft-sampling but combine the sets of data from three transformations into one set. The variant influences the gradient computation on Line 9-10 in the algorithm. We show the result as 'Mixture' in Table~\ref{tab:meta-train}, where we observe that MetaSets outperforms 'Mixture'. The result indicates that separating the different transformations into different tasks is important to learning the knowledge from different tasks while mixing the data from different transformations may disentangle the knowledge.

According to \cite{arjovsky2019invariant}, when minimizing the losses for different tasks, a alternative is to maximizing the maximum of all the losses, which may increase the convergence speed. We compare MetaSets by minimizing all the losses with MetaSets by minimizing the maximum loss of all the tasks ('Maximum Loss') in Table~\ref{tab:meta-train}, we can observe that MetaSets achieves higher accuracy than 'Maximum Loss', which indicates the necessity of minimizing all the task losses. This observation matches the claim in \cite{sagawa2019distributionally} that directly using \textit{distributionally robust optimization} (DRO) \cite{duchi2016statistics} still achieves low worst-group test accuracy.

\begin{table}[ht]
    \centering
    \caption{Time for each epoch (s) and accuracy (\%) for PointNet/MetaSets based on the dynamic transformation and the static transformation.}
    \label{table:static}
    \begin{tabular}{ccccc}
    \toprule
        \multirow{2}{60pt}{Transformation} & \multicolumn{2}{c}{PointNet} &  \multicolumn{2}{c}{MetaSets} \\
        & Time & Acc & Time & Acc \\
        \midrule
        Dynamic & 358 & 63.16 & 1423 & 68.28 \\
        Static & 334 & 45.89 & 1350 & 49.21\\
        \bottomrule
    \end{tabular}
\end{table}

\subsection{Static or Dynamic Transformation}
In MetaSets, we need to randomize the parameters $\vec{v}$, $P_1$ and $P_2$ in every training iteration (dynamic transformation), which needs an extra transformation cost in every iteration. A more efficient way is using static transformation, where we first transform each point cloud with one parameter into a transformed point cloud and form a set of transformed point clouds. We demonstrate that dynamic transformation is necessary and does not introduce two much cost than static transformation. We compare PointNet/MetaSets based on dynamic transformation with PointNet/MetaSets based on static transformation. As shown in Table~\ref{table:static}, the dynamic transformation approach achieves much higher performance than static transformation in accuracy but only little more cost time for each epoch for both PointNet and MetaSets. Also, the number of epochs to converge is about 30 for all experiments. Therefore, the dynamic transformation significantly improves the performance with only little more time cost. 
\begin{figure}[ht]
    \centering
    \subfigure[$\eta$]{\includegraphics[height=.28\textwidth]{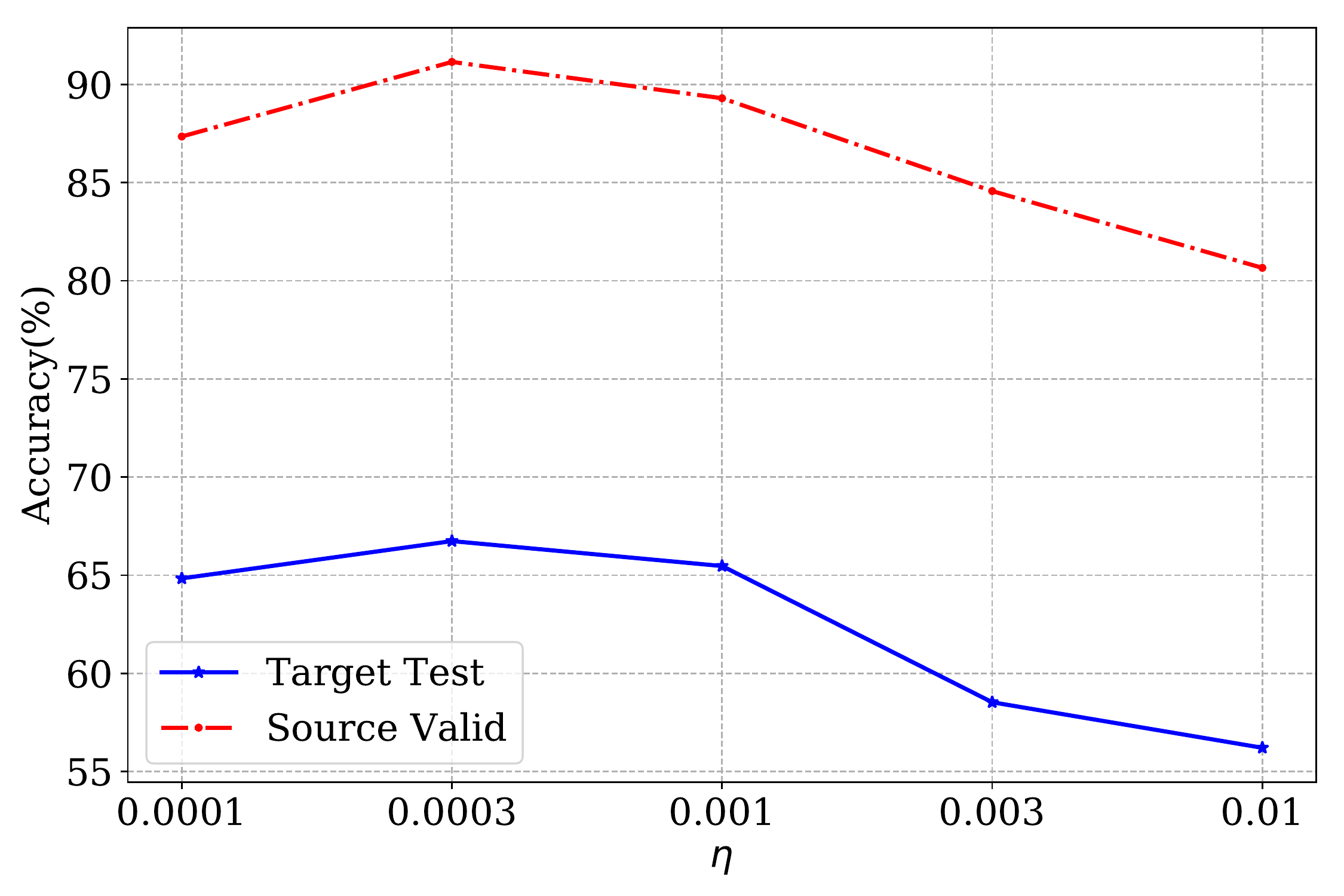}}
    \subfigure[$\beta$]{\includegraphics[height=.28\textwidth]{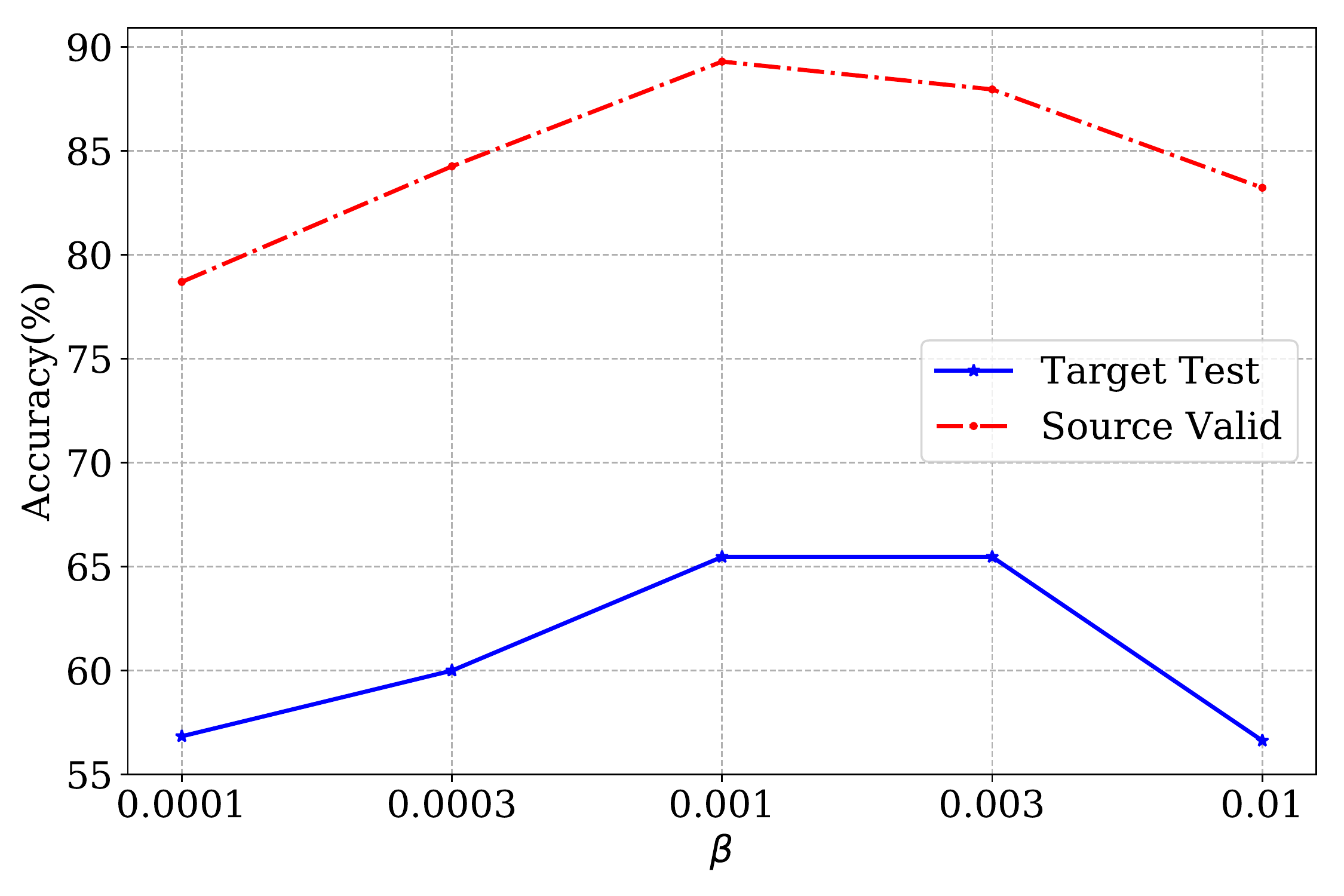}}
    \vspace{-5pt}
    \caption{Sensitivity analyses about $\eta$ and $\beta$ on the ModelNet$\rightarrow$ScanObjNN benchmark.}
    \label{fig:parameter_sensitivity}
\end{figure}
\subsection{Parameter Sensitivity}
We test the classification accuracy for different learning rates $\eta$ and $\beta$ on the ModelNet$\rightarrow$ScanObjNN benchmark. 
To find the best-performing value of $\eta$, we fix the $\beta$, and vice versa. The results are shown in Figure~\ref{fig:parameter_sensitivity}, we can observe that $\eta$ is not sensitive in the range of  $[0.0001,0.001]$ and $\beta$ is not sensitive in the range of $[0.001,0.003]$. However, even the accuracy drops out of these ranges. The trend of the validation accuracy curve and the test accuracy curve are similar, which indicates that the best learning rates can be obtained by cross-validation on the source validation set.

{
\bibliography{cvpr_3ddg.bib}
\bibliographystyle{ieee_fullname}
}
\end{document}